\documentclass[lettersize,journal]{IEEEtran}
\usepackage{amsmath,amsfonts}
\usepackage{algorithmic}
\usepackage{algorithm}
\usepackage{array}
\usepackage[caption=false,font=normalsize,labelfont=sf,textfont=sf]{subfig}
\usepackage{textcomp}
\usepackage{stfloats}
\usepackage{url}
\usepackage{verbatim}
\usepackage{graphicx}
\usepackage{cite}
\usepackage{import}
\usepackage{booktabs} 
\usepackage{makecell}
\usepackage{color,soul}
\begin{document}

\title{mmHand: Toward Pixel-Level-Accuracy Hand Localization Using a Single Commodity mmWave Device}

\author{Xiaoyu Zhang, Zhengxiong Li, Chenhan Xu, Luchuan Song, Huining Li, Hongfei Xue, Yingxiao Wu, Wenyao Xu 
\thanks{Xiaoyu Zhang and Wenyao Xu are with the Department of Computer Science and Engineering, State University of New York at Buffalo, Amherst, NY 14068 USA (email: zhang376@buffalo.edu, wenyaoxu@buffalo.edu).} 
\thanks{Zhengxiong Li is with the Department of Computer Science and Engineering, University of Colorado Denver, Denver, CO 80204 USA (email: zhengxiong.li@ucdenver.edu).} 
\thanks{Chenhan Xu and Huining Li are with the Department of Computer Science, North Carolina State University, Raleigh, NC 27695 USA (email: cxu34@ncsu.edu, huiningl@buffalo.edu).} 
\thanks{Luchuan Song is with the Department of Computer Science, University of Rochester, Rochester, NY 14627 USA (email: lsong11@ur.rochester.edu).} 
\thanks{Hongfei Xue is with the Department of Computer Science, University of North Carolina at Charlotte, Charlotte, NC 28223 USA (email: hongfei.xue@charlotte.edu).}
\thanks{Yingxiao Wu is with the College of Computer Science and Technology, Hangzhou Dianzi University, Hangzhou, Zhejiang 310018, China (email: wuyingxiao@hdu.edu.cn).}
\thanks{Copyright (c) 2025 IEEE. Personal use of this material is permitted. Permission from IEEE must be obtained for all other uses, in any current or future media, including reprinting/republishing this material for advertising or promotional purposes, creating new collective works, for resale or redistribution to servers or lists, or reuse of any copyrighted component of this work in other works.}
\thanks{Published in \emph{IEEE Internet of Things Journal}, vol. 12, no. 12, pp. 20800--20814, 2025. DOI: 10.1109/JIOT.2025.3546560.}
}

\markboth{IEEE INTERNET OF THINGS JOURNAL}%
{Zhang \MakeLowercase{\textit{et al.}}: mmHand: Toward Pixel-Level-Accuracy Hand Localization Using a Single Commodity mmWave Device}


\maketitle

\begin{abstract}
    The hand localization problem has been a longstanding focus due to its many applications. The task involves modeling the hand as a singular point and determining its position within a defined coordinate system. However, due to data modality limitations, existing hand localization technologies face several challenges. For example, vision-based localization raises privacy concerns, while wearable-based methods compromise user comfort. In this paper, we introduce mmHand, a new device-free, privacy-preserving dynamic hand localization system with pixel-level accuracy, using a single commodity mmWave device. We first propose a mmImage generation tool to fully extract spatial information from raw mmWave data and introduce a novel 2D image-format representation of mmWave data. Next, we design a framework that provides a new quality evaluation method and pixel space labeling for the mmWave data. Finally, we present a cross-modality spatial feature-enhanced model with high spatial feature extraction capabilities, which can accurately localize hand positions at the pixel level in the mmWave radar U-V pixel coordinate system. We evaluate the system with experiments on 12 subjects in three scenarios, and the results across four metrics demonstrate the effectiveness of our hand localization system.
\end{abstract}


\section{Introduction}

\IEEEPARstart{H}{and} localization has been widely applied in various fields, including AR/VR, robotics, and gaming. Many companies and organizations are actively developing accurate hand localization methods based on two primary types of human-computer interaction. The first is contact-based hand localization, such as Meta Quest \cite{MetaQuest}, which typically requires a touch controller to determine the hand’s position. However, contact devices are often seen as burdensome to users, compromising the overall experience. The second type is non-contact-based solutions, such as Ultraleap \cite{Ultraleap} and MediaPipe \cite{zhang2020mediapipe}, which use cameras to provide high-resolution tracking in most environments. However, camera footage in the hand localization process raises privacy concerns. In contrast, RF sensors offer a privacy-preserving alternative as non-visual devices. Despite this advantage, most RF sensors face challenges in achieving precise hand localization due to their relatively low resolution.

\begin{figure}[htb]
  \centering
  \includegraphics[width=\linewidth]{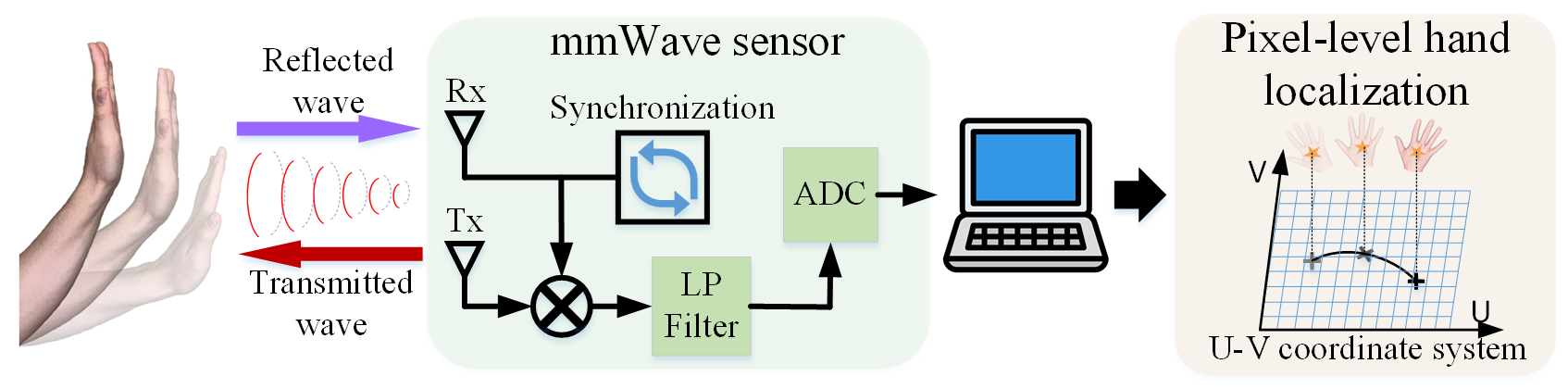}
  \caption{The mmWave sensor receives signals reflected from the hand and processes them for dynamic hand localization.}
  \label{fig:mmHand}
\end{figure}

Given the current challenges, mmWave radar, a type of RF sensor, offers several advantages for hand localization products: (1) \textbf{High resolution: } operating in the GHz band, it provides millimeter-level accuracy; (2) \textbf{Fast response: } it can respond as quickly as a 500 FPS camera, allowing for the localization of fast-moving hands; (3) \textbf{Privacy-preserving: } it doesn’t rely on vision-based signals; (4) \textbf{Integratable: } its compact size enables easy integration into mobile devices for various applications; (5) \textbf{Device-free: } utilizing frequency modulated continuous wave (FMCW) technology, it detects and localizes hands without requiring physical contact with the device.

There are several approaches to hand localization using mmWave sensors \cite{regani2021mmwrite, ninos2022radar}. However, existing methods struggle to effectively extract the full information from mmWave data and still rely on more complex and expensive mmWave devices to localize the hand.

In this paper, we propose mmHand to achieve pixel-level accuracy in hand localization. As shown in Figure \ref{fig:mmHand}, the mmHand system utilizes the mmWave sensor to emit signals and capture their reflections from the hand, which are then processed for localization. The development of our system presented three key challenges: (1) The raw mmWave data are time-sequence energy signals, which are fundamentally different from image data, making it difficult to associate the time-series signal with pixels. To address this, we develop a mmImage generation tool to convert raw mmWave data into a new image format, correlating it with pixels. (2) Existing methods cannot label the image-formatted mmWave data for hand localization model training, so we design a cross-modality-based labeling method to efficiently generate accurate pixel-space labels. (3) Current hand localization models lack sufficient ability to extract high-quality spatial features from the image-formatted mmWave data. To enhance feature extraction, we connect the mmWave data with higher-resolution depth images, leveraging the spatial features of the depth images to guide mmWave hand localization model training and improve feature extraction. Extensive experiments demonstrate that the mmHand system performs well in hand localization, and the enhanced model effectively extracts high-quality spatial features.

In conclusion, our contributions are listed as follows:
\begin{itemize}
\item We present a tool to generate a new image-format representation of mmWave data, called mmImage. This tool fully extracts the spatial information from the mmWave data and associates the time-series data with the pixel concept.
\item We design a new evaluation method for mmWave data quality that focuses on spatial information and effectively identifies low-quality data with indistinct spatial details caused by interference.
\item Leveraging the data format similarity between the mmImage and synchronized depth images, we propose a novel mmWave data labeling method that efficiently provides accurate pixel-space labels for training image-based task models.
\item We propose a cross-modality spatial-feature-enhanced hand detection model based on the mmImage. The model utilizes the spatial feature similarity between the mmImage and depth image to build an efficient cross-modality connection, enhancing the spatial feature extraction of the mmImage under the guidance of the higher-resolution depth image. Additionally, our spatial-feature-enhanced model can facilitate effective interaction between all image-format data modalities with similar spatial features.
\item We design an end-to-end hand localization system using a single commodity mmWave device. It requires no specific sensor placement and can localize hand positions from single-frame mmWave data. Extensive experiments involving 12 subjects across three scenarios, using four metrics, demonstrate the effectiveness of our proposed hand localization system.
\end{itemize}

\section{Background}

\subsection{Depth Image Modality}

The depth camera is a widely used sensor capable of accurately measuring the distance between an object and the camera. Currently, three depth sensing technologies are commonly employed in the depth cameras, which are outlined below.
\begin{itemize}
    \item \textbf{Structured light (e.g., Kinect V1 and RealSense): } This type of depth camera emits invisible infrared lasers at a specific wavelength and captures depth information by analyzing distortions in the reflected encoded pattern.
    \item \textbf{Stereo vision (e.g., Stereo IR 170 and ZED 2): } This depth camera captures two images of an object from different viewpoints and calculates depth by measuring the disparity between corresponding points in the two images.
    \item \textbf{Time of Flight (e.g., Kinect V2 and SoftKinect): } This type of depth camera continuously emits laser pulses toward an object and measures depth by calculating the round-trip time of the reflected pulses.
\end{itemize}

Typically, the data format of a depth image is $(\mathcal{W}, \mathcal{H})$, where $\mathcal{W}$ and $\mathcal{H}$ represent the width and height of the image, respectively.

\begin{figure}[htb]
  \centering
  \includegraphics[width=\linewidth]{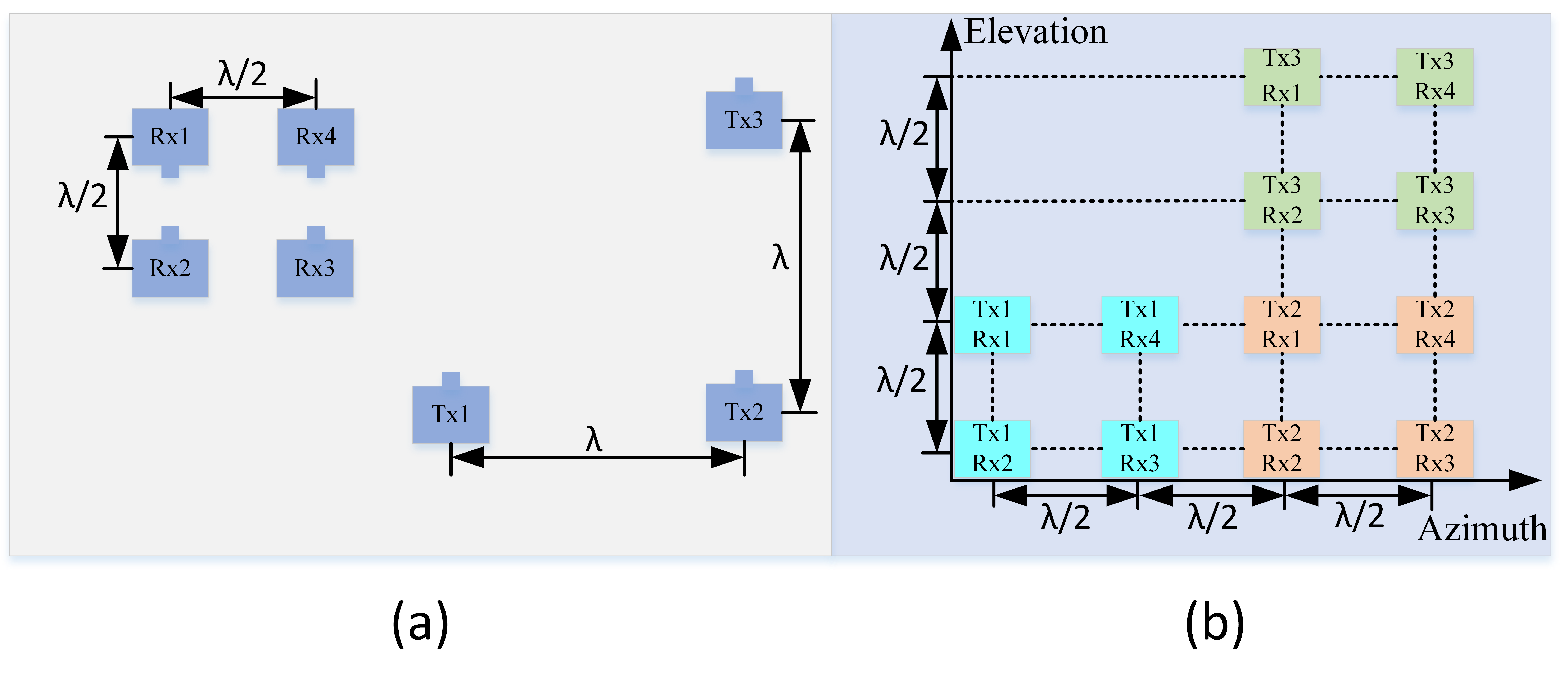}
  \caption{An example of a mmWave sensor with 3 transmitters (Tx) and 4 receivers (Rx) is shown. On the left is the antenna placement and spacing, while on the right is the corresponding virtual antenna array.}
  \label{fig:mmWave}
\end{figure}

\subsection{mmWave Data Modality} \label{sec:modality}

Figure \ref{fig:mmWave} illustrates an example of a mmWave sensor with 3 transmitting antennas (Tx) and 4 receiving antennas (Rx). During each frame period, the mmWave sensor samples $C$ chirps of data using MIMO (Multiple-Input Multiple-Output) technology, with each Tx sequentially emitting signals that are received by all Rx in one chirp. The mmWave sensor uniformly samples $N$ signal points from each received Tx signal by each Rx. Consequently, the format for one frame of mmWave data is represented as $(\mathcal{T}, \mathcal{R}, \mathcal{C}, \mathcal{N})$, where $\mathcal{T}$, $\mathcal{R}$, $\mathcal{C}$, and $\mathcal{N}$ denote the number of Tx, Rx, chirps, and samples, respectively.

Due to its high sampling rate, the mmWave sensor achieves high-frequency resolution. Using FMCW technology, differences in frequency are converted into distance information, so high-frequency resolution contributes to the sensor's range resolution. Furthermore, the presence of the 2D virtual antenna array shown in Figure \ref{fig:mmWave} allows the mmWave radar to provide a 2D spatial view for detecting object distances, akin to a depth camera. This capability makes it feasible to establish a cross-modality connection between mmWave data and depth images, as both modalities share spatial information characteristics. Details of the interaction between these two modalities are presented in Section \ref{sec:crossmodality}.

\section{mmHand system framework}

\begin{figure*}[htb]
  \centering
  \includegraphics[width=\linewidth]{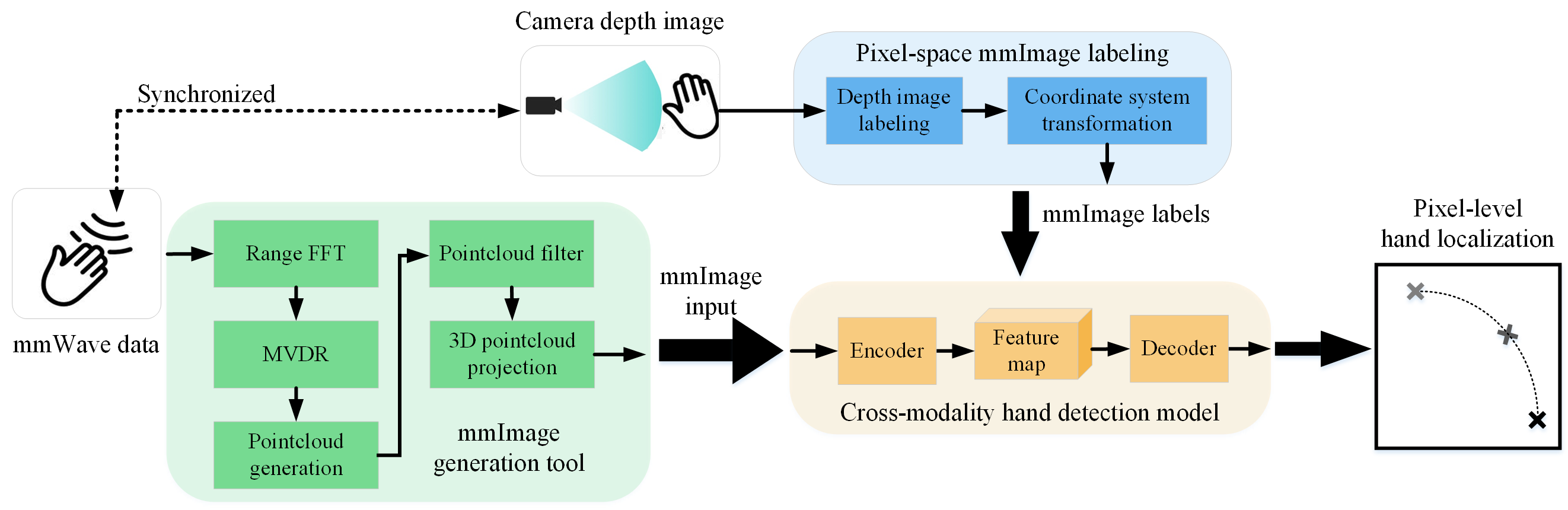}
  \caption{The mmHand system framework.}
  \label{fig:framework}
\end{figure*}

Figure \ref{fig:framework} provides an overview of the mmWave-based hand localization system with pixel-level accuracy. The mmHand system consists of four components: the mmImage generation tool, pixel-space mmImage labeling and cross-modality hand detection model, and pixel-level hand localization. First, we introduce mmImage, a novel representation of mmWave data that fully captures spatial information and converts it into an image format. Second, we leverage synchronized depth camera images to generate accurate pixel-space hand location labels for the mmImage. Third, we establish a cross-modality connection between mmImages and depth images, enhancing spatial feature extraction for improved hand detection performance. Finally, we develop a context-aware algorithm that mitigates inaccurate predictions, ensuring precise and robust dynamic hand localization.

\section{mmImage Generation Tool}
\begin{figure*}[htb]
  \centering
  \includegraphics[width=\linewidth]{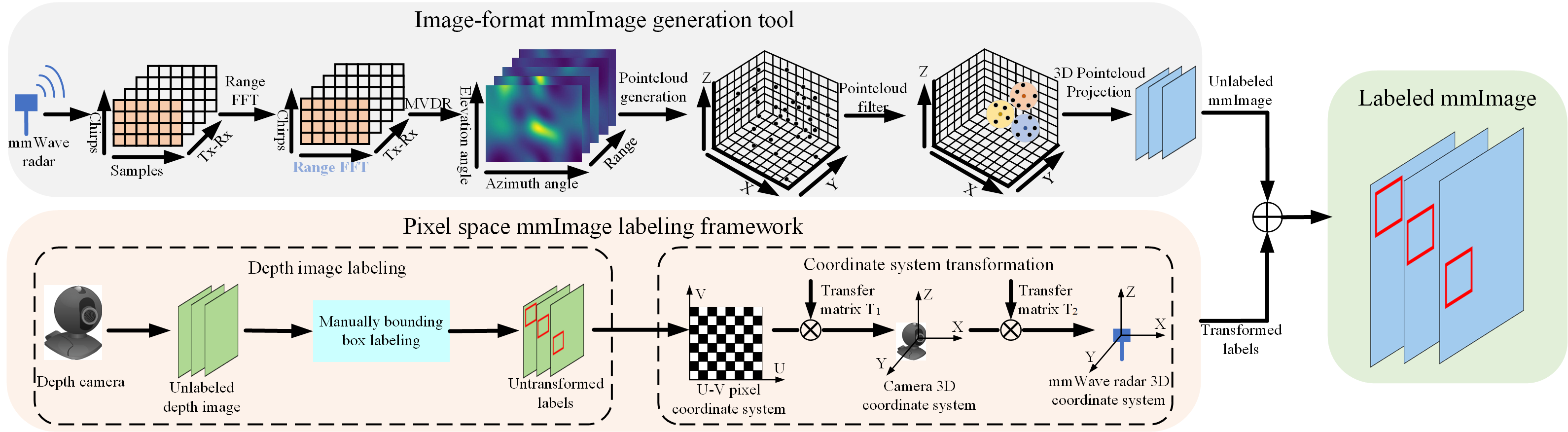}
  \caption{The mmImage generation tool for image-format conversion and the pixel space mmImage labeling framework.}
  \label{fig:Label}
\end{figure*}

In this section, we introduce the mmImage generation tool, which converts a frame of raw time-series mmWave data $(\mathcal{T}, \mathcal{R}, \mathcal{C}, \mathcal{N})$ into a new image-format representation called mmImage $(\mathcal{W}, \mathcal{H})$. As illustrated in Figure \ref{fig:Label}, the mmImage generation process consists of five steps: range FFT, MVDR data processing, point cloud generation, point cloud filtering, and 3D point cloud projection. We will describe each of these steps in detail.

\subsection{Range FFT}

When the mmWave sensor transmits an FMCW signal, it propagates to multiple objects at different positions, resulting in varying time delays in the received reflections. The range FFT processes these time delays by converting them into frequency peaks, effectively revealing the distance differences between objects. After applying the range FFT, the raw mmWave data $(\mathcal{T}, \mathcal{R}, \mathcal{C}, \mathcal{N})$ is transformed into the range bin format $(\mathcal{T}, \mathcal{R}, \mathcal{C}, \mathcal{D})$, where $\mathcal{D}$ retains the same value as $\mathcal{N}$ and represents the range dimension.

\subsection{Minimum Variance Distortionless Response (MVDR)}

As shown in Figure \ref{fig:mmWave}(a), the mmWave sensor is equipped with multiple transmitting and receiving antennas. Using MIMO (multiple-input, multiple-output) technology, different transmitting antennas operate in separate time slots, allowing the receiving antennas to form a virtual antenna array, as illustrated in Figure \ref{fig:mmWave}(b). The two-dimensional spatial structure of this antenna array enables the mmWave sensor to detect objects in both the azimuth and elevation dimensions. 

MVDR, also known as Capon Beamforming, is a valuable tool in mmWave data processing. After obtaining the range bins from the Range FFT step, we apply MVDR to combine the azimuth and elevation information, generating range-azimuth-elevation data $(\mathcal{D}, \Theta, \Phi)$ in the spherical coordinate system, where $\Theta$ and $\Phi$ represent the azimuth and elevation angles, respectively.

MVDR is a signal processing technique that enhances signals from the desired direction while suppressing interference from others. In 3D point cloud generation, MVDR enables precise beamforming by dynamically adjusting the antenna array to focus on specific directions. By capturing signals from multiple angles, the system accumulates a dense set of data points, resulting in a more detailed and refined 3D point cloud. This enhanced data density allows the MVDR beamformer to generate a comprehensive and high-resolution representation of hand regions and contours.

\subsection{Pointcloud Generation}

Compared to the spherical coordinate system, the 3D Cartesian coordinate system offers better spatial representation and can more directly express an object's spatial contour and precise location. Additionally, the 3D Cartesian coordinate system has the advantage of being more easily associated with the U-V pixel coordinate system, facilitating pixel-level hand localization. Therefore, we convert the $\mathcal{D} \cdot \Theta \cdot \Phi$ points from the range-azimuth-elevation data $(\mathcal{D}, \Theta, \Phi)$ into the pointcloud $\mathcal{P} = (\mathcal{D} \cdot \Theta \cdot \Phi, (x, y, z, e))$ using Equation (\ref{equ:Coordinator}). In this representation, the second dimension consists of the point's 3D coordinates $(x, y, z)$ in the 3D Cartesian coordinate system and the corresponding energy $e$, with the origin of the system located at the center of the mmWave sensor antenna array. However, many of the points are not related to the object, so in the next step, we apply a point cloud filter to refine the generated point cloud and retain only the object-related points.
\begin{equation}
\small
\left\{
\begin{aligned}
  &x=r\cos \phi \cos \theta \\
  &y=r\cos \phi \sin \theta   \qquad r \in \mathcal{D}, \theta \in \Theta , \phi \in <\Phi\\
  &z=r\sin \phi 
  \label{equ:Coordinator}
\end{aligned}
\right.
\end{equation}

\subsection{Pointcloud Filter}

Due to signal attenuation, the received signal loses energy as it propagates over longer distances. In hand localization, since the hand is typically positioned close to the mmWave sensor, points corresponding to the hand generally exhibit higher energy. However, due to the resolution limitations of the sensor, high-energy points from different objects—such as the human trunk and hand—may become indistinguishably mixed, leading to challenges in accurate segmentation and localization.

\begin{algorithm}
    \renewcommand{\algorithmicrequire}{\textbf{Input:}}
    \renewcommand{\algorithmicensure}{\textbf{Output:}}
    \caption{Pointcloud clustering algorithm.}
    \label{Alg:Clustering}
    \begin{algorithmic}[1]
        \REQUIRE unprocessed pointcloud set contains the 3D coordinate and energy value $\mathcal{P}=(\mathcal{D}\cdot \Theta \cdot \Phi, (x, y, z, e))$, cluster center number $\mathcal{V}$, cluster size $\mathcal{S}$
        \ENSURE processed pointcloud set $\mathcal{G}$
        \STATE sort $\mathcal{P}$ in descending order based on the energy $e$
        \FOR {$i=1, ..., \mathcal{V}$}
        \STATE choose the first point $p_i$ in $\mathcal{P}$
        \STATE calculate the distance of all the points to $p_i$ based on the 3D coordinate $(x, y, z)$ and get a distance set $D_i$
        \STATE sort $D_i$ in ascending order
        \STATE $D_i\longleftarrow D_i(1:\mathcal{S})$
        \STATE $\mathcal{G}\longleftarrow \mathcal{G}\cup D_i$
        \STATE remove $D_i$ from $\mathcal{P}$
        \ENDFOR
        \STATE \textbf{return} $\mathcal{G}$
    \end{algorithmic}
\end{algorithm}

To distinguish points from different objects, we use a clustering algorithm to filter out points with high discrimination. As outlined in Algorithm \ref{Alg:Clustering}, we first sort the unprocessed pointcloud $\mathcal{P}$ in descending order based on the energy $e$. For each cycle $0 < i < \mathcal{V}$, the point with the highest energy is set as the cluster center $p_i$. In steps 4-6, we find the $\mathcal{S}$ nearest points to the cluster center $p_i$ (including $p_i$ itself), and these $\mathcal{S}$ points form a cluster with $p_i$ as the center. In steps 7-8, we add the $\mathcal{S}$ points to the processed point cloud set $\mathcal{G}$ and remove them from the unprocessed point cloud set $\mathcal{P}$ to ensure they are not considered in subsequent clustering cycles. After the clustering algorithm, we obtain a purified pointcloud $\mathcal{G} = (\mathcal{V} \cdot \mathcal{S}, (x, y, z, e))$ with $\mathcal{V} \cdot \mathcal{S}$ points.

\subsection{3D Pointcloud Projection}

As discussed in Section \ref{sec:modality}, due to the spatial information similarity between mmWave data and depth images, we aim to establish a cross-modality connection between these two data modalities to improve hand localization. Since the 3D point cloud differs from the 2D depth image in data format, it is important to minimize the modality gap for an effective cross-modality connection. As shown in Figure \ref{fig:mmWave}, the mmWave sensor’s virtual antenna array provides a 2D view on the antenna array plane, with the vertical axis representing the depth value. Following the data generation method used by depth cameras, we project the 3D pointcloud onto the antenna array plane, creating a 2D image-format representation of the mmWave data in the sensor's coordinate system. Finally, we discretize the data points in the image and crop it to a fixed size $(\mathcal{W}, \mathcal{H})$, which we refer to as mmImage.

\begin{table*}
\small
  \centering
  \caption{Overview of mmImage Generation.}
  \label{tab:mmImage}
  \begin{tabular}{cccccc}
    \toprule
    Data type & Next step & Data format\\
    \midrule
    Raw mmWave data & Range FFT & $(\mathcal{T}, \mathcal{R}, \mathcal{C}, \mathcal{N})$\\
    Range bin & MVDR & $(\mathcal{T}, \mathcal{R}, \mathcal{C}, \mathcal{D})$\\
    Range-azimuth-elevation data & Pointcloud generation & $(\mathcal{D}, \Theta, \Phi)$\\
    Pointcloud & Pointcloud filter & $(\mathcal{D}\cdot \Theta \cdot \Phi, (x, y, z, e))$\\
    Filtered pointcloud & 3D pointcloud projection & $(\mathcal{V}\cdot \mathcal{S}, (x, y, z, e))$\\
    mmImage & $\setminus $ & $(\mathcal{W}, \mathcal{H})$\\
  \bottomrule
\end{tabular}
\end{table*}

\begin{table}
\small
  \centering
  \caption{Symbol Definition and Description.}
  \label{tab:symbol}
  \begin{tabular}{cl}
    \toprule
    Symbol & Definition description \\
    \midrule
    $\mathcal{T}$ & The number of the transmitting antennas\\
    $\mathcal{R}$ & The number of the receiving antennas\\
    $\mathcal{C}$ & The number of chirps per frame\\
    $\mathcal{N}$ & The number of samples per chirp\\
    $\mathcal{D}$ & The number of ranges\\
    $\Theta$ & The number of azimuth angles\\
    $\Phi$ & The number of elevation angles\\
    $\mathcal{V}$ & The number of clusters per frame\\
    $\mathcal{S}$ & The number of points per cluster\\
    $\mathcal{W}$ & The width of mmImage\\
    $\mathcal{H}$ & The height of mmImage\\
  \bottomrule
\end{tabular}
\end{table}

\begin{figure*}[htb]
  \centering
  \includegraphics[width=\linewidth]{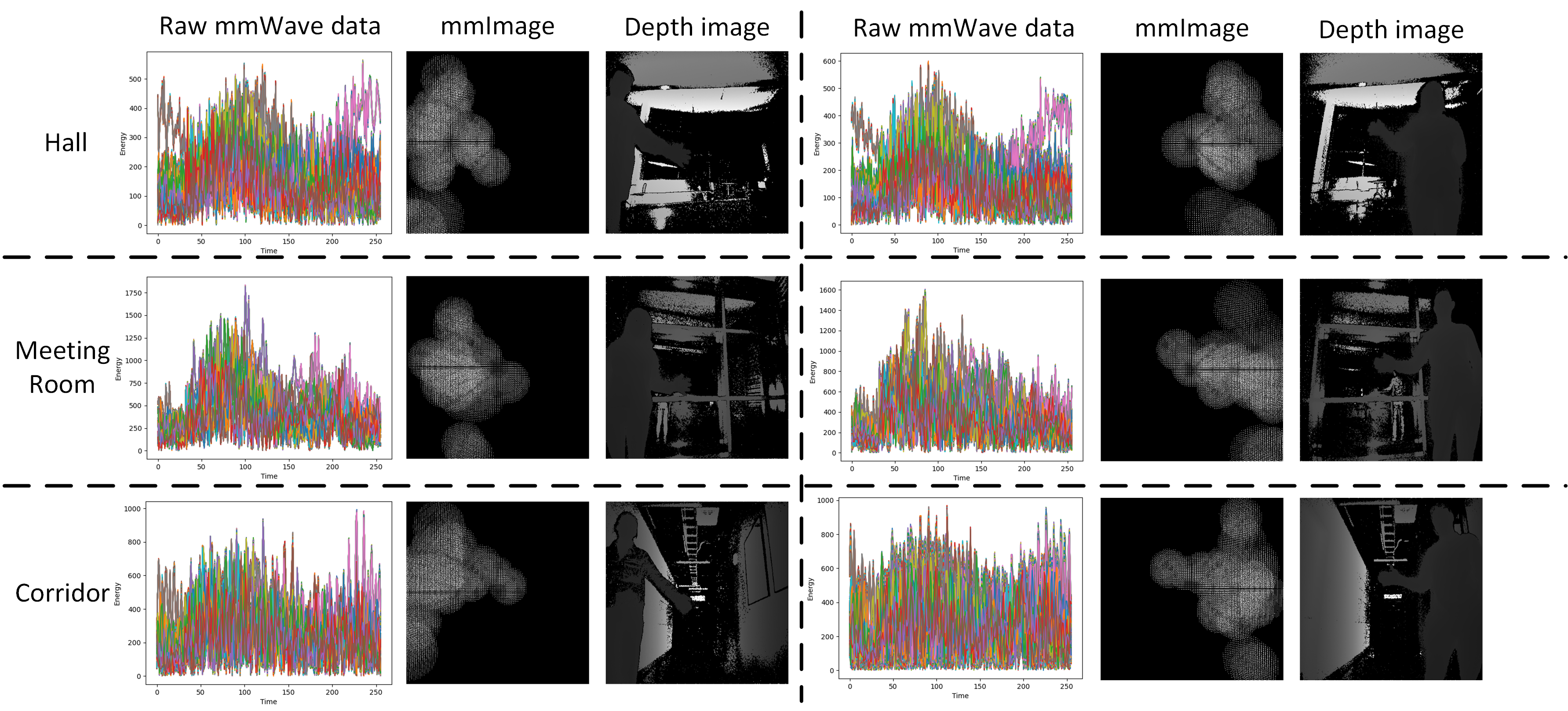}
  \caption{Examples of the raw mmWave data, generated mmImage and synchronized depth image in three scenarios (Hall, meeting room and corridor).}
  \label{fig:mmImage_example}
\end{figure*}

Table \ref{tab:mmImage} details the steps involved in mmImage generation along with the corresponding data format at each stage, while Table \ref{tab:symbol} defines the symbols used in Table \ref{tab:mmImage}. Figure \ref{fig:mmImage_example} presents a comparative analysis of the raw mmWave data, the generated mmImage, and the corresponding depth image across three distinct environments: a hall, a meeting room, and a corridor. Each set of images corresponds to a different subject. As illustrated in Figure \ref{fig:mmImage_example}, our mmImage generation tool effectively extracts spatial information from raw mmWave data and maps it into pixel space, producing an mmImage that closely aligns with the depth image captured by the depth camera.

\section{mmImage-based Hand Localization} \label{sec:crossmodality}

In this section, we introduce an mmImage-based hand localization system that fully leverages the spatial information similarity between the mmImage and the synchronized depth image. This system enables a new quality evaluation method and a 2D pixel space labeling approach for time-series mmWave data, delivering high-accuracy, pixel-level hand localization.

\subsection{mmWave Data Quality Evaluation}

Due to the sensitivity of mmWave signals to environmental interference and internal circuit deviations, the sampled raw mmWave data can vary significantly, leading to differences in the quality of the generated mmImages. To effectively evaluate the mmWave data and filter out poor-quality data, we propose a novel mmWave data quality evaluation method based on the structural similarity index measure (SSIM).
\begin{equation}
\small
  SSIM(x,y)=\frac{(2\mu _{x} \mu _{y} +(k_{1} L)^{2} )(2\sigma _{xy} +(k_{2} L)^{2} ))}{(\mu _{x}^{2} +\mu _{y}^{2} +(k_{1} L)^{2} )(\sigma _{x}^{2} +\sigma _{y}^{2} +(k_{2} L)^{2} )}.
  \label{equ:SSIM}
\end{equation}
The details of the SSIM calculation are based on Equation (\ref{equ:SSIM}), where $x$ and $y$ are the two images being compared to calculate the SSIM value. $\mu_{x}$ and $\mu_{y}$ represent the pixel sample means of the two images, $\sigma_{x}^{2}$ and $\sigma_{y}^{2}$ are the variances, and $\sigma_{xy}$ is the covariance. Additionally, $L$ represents the dynamic range of pixel values, and $k_{1}$ and $k_{2}$ are two constants, typically set to $k_{1} = 0.01$ and $k_{2} = 0.03$ by default. In situations where mmWave data are collected under poor conditions with significant interference, object signals may be obscured by environmental noise. In these cases, the mmImage generation tool may struggle to effectively extract spatial information from the noisy mmWave data, resulting in a low similarity between the generated mmImage and the synchronized depth image. This is captured by the SSIM metric. We compute the SSIM value for all mmImages against their corresponding synchronized depth images and classify mmImages with SSIM values above a predefined threshold as high-quality data.

\subsection{Pixel Space mmImage Labeling} \label{sec:label}

Due to the inherent representation of mmWave data as a time-series energy signal, the signal at each time point is a mixture of reflections from both the object and the environment, making it impossible to label the object in the raw mmWave data. As shown in Figure \ref{fig:Label}, we propose a novel pixel-space mmImage labeling method that quickly provides accurate object labels for the mmWave data. This method can be broadly applied to image-based tasks such as object detection and localization. The detailed labeling steps are as follows:

\noindent
(a) First, the mmWave sensor and depth camera are positioned at a fixed relative distance to collect data.

\noindent
(b) Second, the mmWave data and depth images are synchronized frame by frame using the timestamp.

\noindent
(c) Third, a pre-trained object detection model is used to detect the object in pixel space from the collected depth images. The pre-trained model can either be obtained by training on a manually labeled depth image dataset or by using a trained model from existing object detection works.

\noindent
(d) Fourth, since the detected labels are in the depth camera's U-V pixel coordinate system and the mmWave data are in the mmWave sensor's coordinate system, the detected labels must be transformed by multiplying them with two transfer matrices ($T_{1}$ and $T_{2}$) to align the coordinate systems.

\noindent
(e) Finally, each frame of the transformed label is used as the object label for the synchronized mmImage generated from the raw mmWave data.

Since the new mmWave data labeling method is based on the synchronized depth image, it is independent of different mmWave data processing methods. The primary factor affecting labeling accuracy is the synchronization deviation. The transfer matrix $T_{1}$, which converts the depth camera's U-V pixel coordinate system to its 3D coordinate system, depends on the camera's internal fixed parameters. Similarly, the transfer matrix $T_{2}$, which converts from the depth camera's 3D coordinate system to the mmWave sensor's 3D coordinate system, depends on the relative distance between the two sensors. In the experiment, we maintain a fixed relative distance by placing the two sensors in specific locations on a wooden board. As a result, both transfer matrices are linear transformations with fixed parameters. We assume that these linear transformations can be learned by the deep learning model during training, enabling us to simplify the use of untransformed labels in the experimental evaluation.

\subsection{Hand Detection Module}

After obtaining the labeled mmImage data, a deep learning model is required to detect the hand from the mmImage. Due to its fast response and strong performance in object detection \cite{redmon2018yolov3}, we use the Yolov3 model as our hand detection model structure. However, despite the mmImage generation tool effectively extracting spatial information from the raw mmWave data, the inherent lower resolution of the mmWave radar compared to cameras results in the mmImage having lower data quality than a depth image. This limitation hinders the mmImage-based model’s ability to extract effective spatial features for hand detection. As shown in Figure \ref{fig:yolo}, we propose a cross-modality spatial-feature enhanced model, which establishes a guidance interaction from the depth-image-based model to the mmImage-based model. The depth-image-based model, a pre-trained Yolov3 model with high hand detection accuracy, includes a feature encoder to extract effective spatial features from the depth image. By introducing a cross-modality loss function between the feature maps of the mmImage-based encoder and the depth-image-based encoder, the mmImage-based model is guided to extract more effective spatial features from the mmImage, leading to higher accuracy in hand detection.

\subsubsection{Cross-modality Loss} \label{sec:cross-loss}

To enhance the feature extraction capability of the mmImage-based encoder, it is reasonable to constrain it to extract feature maps of similar quality to those produced by the depth-image-based encoder, considering their spatial information similarity. Therefore, we design a cross-modality loss function based on Cosine Similarity to guide the training of the mmImage-based hand detection model.

After feeding the mmImage and synchronized depth image into the mmImage-based encoder and the pre-trained depth-image-based encoder, both encoders extract feature maps $f^{j} =R^{K \cdot W \cdot H } , j=1,2$, where $K$, $W$, and $H$ represent the number of channels, width, and height of the feature map, and $j=1$ and $j=2$ correspond to the feature maps of the mmImage-based encoder and the depth-image-based encoder, respectively. Next, we reshape the feature map $f^{j}$ into sets of row vectors $\mathcal{F}_{r}^{j}=(\overrightarrow{v_{1}^{j} } ,\overrightarrow{v_{2}^{j} } ,\dots ,\overrightarrow{v_{K \cdot H }^{j} })$ and column vectors $\mathcal{F}_{c}^{j}=(\overrightarrow{v_{1}^{j} } ,\overrightarrow{v_{2}^{j} } ,\dots ,\overrightarrow{v_{K \cdot W }^{j} })$ based on the width dimension $W$ and height dimension $H$, respectively. We then calculate the overall feature map Cosine Similarity using Equation (\ref{equ:Cosine}) to measure the spatial feature similarity between the two feature maps.
\begin{equation}
\small
\begin{aligned}
  &similarity=\frac{cos(\mathcal{F}_{r}^{j=1}, \mathcal{F}_{r}^{j=2} )+cos(\mathcal{F}_{c}^{j=1}, \mathcal{F}_{c}^{j=2} )}{2}, \\
  &cos(\mathcal{F}_{r}^{j=1}, \mathcal{F}_{r}^{j=2} )=\frac{\Sigma _{x=1}^{K \cdot H} \frac{\overrightarrow{v_{x}^{j=1} }\cdot \overrightarrow{v_{x}^{j=2} }}{\left \| \overrightarrow{v_{x}^{j=1} } \right \|_{2} \left \| \overrightarrow{v_{x}^{j=2} }\right \|_{2}}}{K \cdot H}, \\
  &cos(\mathcal{F}_{c}^{j=1}, \mathcal{F}_{c}^{j=2} )=\frac{\Sigma _{x=1}^{K \cdot W} \frac{\overrightarrow{v_{x}^{j=1} }\cdot \overrightarrow{v_{x}^{j=2} }}{\left \| \overrightarrow{v_{x}^{j=1} } \right \|_{2} \left \| \overrightarrow{v_{x}^{j=2} }\right \|_{2}}}{K \cdot W}. \\
  \label{equ:Cosine}
\end{aligned}
\end{equation}
Finally, the cross-modality loss $\mathcal{L}_{cross}$ is calculated using Equation (\ref{equ:crossloss}).
\begin{equation}
\small
    \mathcal{L}_{cross}=1-similarity.
    \label{equ:crossloss}
\end{equation}

\subsubsection{Model Loss}

In addition to using the cross-modality loss $\mathcal{L}_{cross}$ to enhance feature extraction, we apply another loss function, $\mathcal{L}_{label}$, during the training of the mmImage-based model for accurate hand detection. The model output includes the location of a bounding box around the hand object and a confidence value. The label loss $\mathcal{L}_{label}$ consists of the bounding box location loss $\mathcal{L}_{bbox}$ and the confidence loss $\mathcal{L}_{conf}$, as defined in Equation (\ref{equ:labelloss}):
\begin{equation}
\small
    \mathcal{L}_{label}=\alpha _{bbox}\cdot \mathcal{L}_{bbox}+\alpha _{conf}\cdot \mathcal{L}_{conf},
    \label{equ:labelloss}
\end{equation}while $\alpha _{bbox}$ and $\alpha _{conf}$ represent the weights of $\mathcal{L}_{bbox}$ and $\mathcal{L}_{conf}$, respectively. The details of the label loss function can be found in \cite{redmon2016you}. Thus, the total model loss function $\mathcal{L}$ is the weighted sum of the two loss functions, as shown in Equation (\ref{equ:loss}).
\begin{equation}
\small
    \mathcal{L}=\alpha _{cross}\cdot  \mathcal{L}_{cross}+\alpha _{label}\cdot  \mathcal{L}_{label},
    \label{equ:loss}
\end{equation}while $\alpha _{cross}$ and $\alpha _{label}$ represent the weights of $\mathcal{L}_{cross}$ and $\mathcal{L}_{label}$, respectively.

\begin{figure}[htb]
  \centering
  \includegraphics[width=\linewidth]{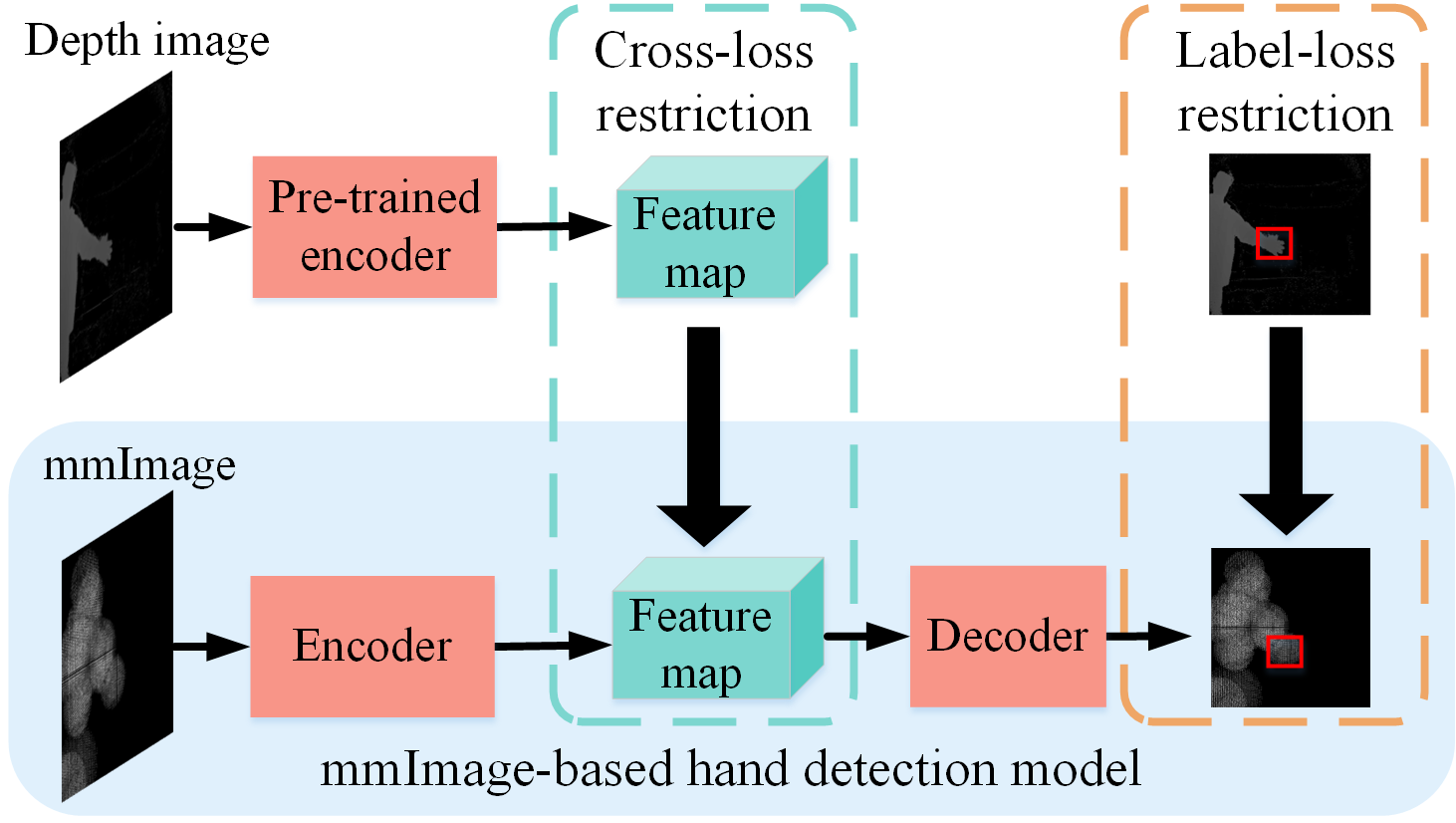}
  \caption{The cross-modality spatial-feature enhanced hand detection model.}
  \label{fig:yolo}
\end{figure}

\begin{algorithm}
    \renewcommand{\algorithmicrequire}{\textbf{Input:}}
    \renewcommand{\algorithmicensure}{\textbf{Output:}}
    \caption{Context-based dynamic hand localization algorithm.}
    \label{alg:tracking}
    \begin{algorithmic}[1]
        \REQUIRE $\mathcal{U}$ consecutive frames of predicted hand locations $A_{1} ,A_{2} ,\dots ,A_{\mathcal{U}} $ and corresponding confidences $conf_{1} ,conf_{2} ,\dots ,conf_{U} $, the combined frame number $\mathcal{W}$
        \ENSURE $\mathcal{U}$ consecutive frames hand locations $\mathcal{O}=B_{1} ,B_{2} ,\dots ,B_{\mathcal{U}}$
        \STATE initializes the output set $\mathcal{O}=\emptyset $
        \FOR {$i=1, ..., \mathcal{U}$}
        \STATE $B_{i}=\frac{A_{i}\cdot conf_{i}+\Sigma_{x=j}^{i-1}B_{x}\cdot conf_{x} \cdot \frac{\mathcal{W}-i+x}{\mathcal{W}} }{conf_{i}+\Sigma_{x=j}^{i-1} conf_{x} \cdot \frac{\mathcal{W}-i+x}{\mathcal{W}} } ,j=max(1,i-\mathcal{W})$
        \STATE add $B_{i}$ in $\mathcal{O}$
        \ENDFOR
        \STATE \textbf{return} $\mathcal{O}$
    \end{algorithmic}
\end{algorithm}

\subsection{Pixel-Level Dynamic Hand Localization}

In hand localization, the hand is typically represented by its center, so pixel-level dynamic hand localization involves accurately identifying the hand center in pixel space. Since the bounding box is a rectangular box that tightly encloses the object, the center of the hand bounding box can be used as the hand center. By applying the trained mmImage-based hand detection model, we can localize the hand in pixel space with a confidence value. However, environmental disturbances and internal circuit deviations can reduce the quality of the mmWave data, making some frames unsuitable for hand localization. Additionally, due to model accuracy limitations, the trained mmImage-based model may fail to accurately localize the hand in certain frames, even when the data quality is good.

To address the challenges mentioned above, we designed a context-based algorithm for accurate dynamic hand localization, leveraging the continuity of hand movement. The details of the dynamic hand localization algorithm are provided in Algorithm \ref{alg:tracking}. First, the mmImage-based hand detection model predicts the hand bounding box for $\mathcal{U}$ consecutive frames, where the center of each bounding box is set as the hand center. The $\mathcal{U}$ predicted hand centers and their corresponding confidence values are used as input for the algorithm. In step 1, we initialize an empty set to store the output hand centers. In steps 2-4, we calculate the hand center for each frame using a weighted summation. Specifically, for the $i_{th}$ frame, we compute a weighted sum of the $i_{th}$ frame's predicted hand center location along with the output hand center locations from the preceding $\mathcal{W}$ frames (if fewer than $\mathcal{W}$ frames are available, we sum all previous $i$ frames). We then divide this sum by the total weight of all combined frames to obtain the weighted average hand center location for the $i_{th}$ frame. Additionally, each preceding frame's weight in the $i_{th}$ frame's localization is determined by the product of its confidence and the time difference relative to the current frame, meaning the hand center location of the current frame depends more heavily on the most recent high-confidence frames. Finally, the algorithm outputs the hand center locations for the $\mathcal{U}$ consecutive frames.

In summary, the mmHand system is comprised of three key components: the mmImage generation tool, cross-modality hand detection, and pixel-level dynamic hand localization. Initially, the mmImage generation tool captures spatial information from raw mmWave data and converts it into pixel space. The generated mmImages are then used to train a hand detection model that benefits from improved feature extraction. Lastly, a context-aware algorithm is employed to overcome challenges related to inconsistent data quality and model detection bias, ultimately achieving precise, pixel-level dynamic hand localization.
\section{Evaluation}

\subsection{Experiment Preparation}

\subsubsection{TestBeds}

\begin{figure}[htb]
  \centering
  \includegraphics[width=\linewidth]{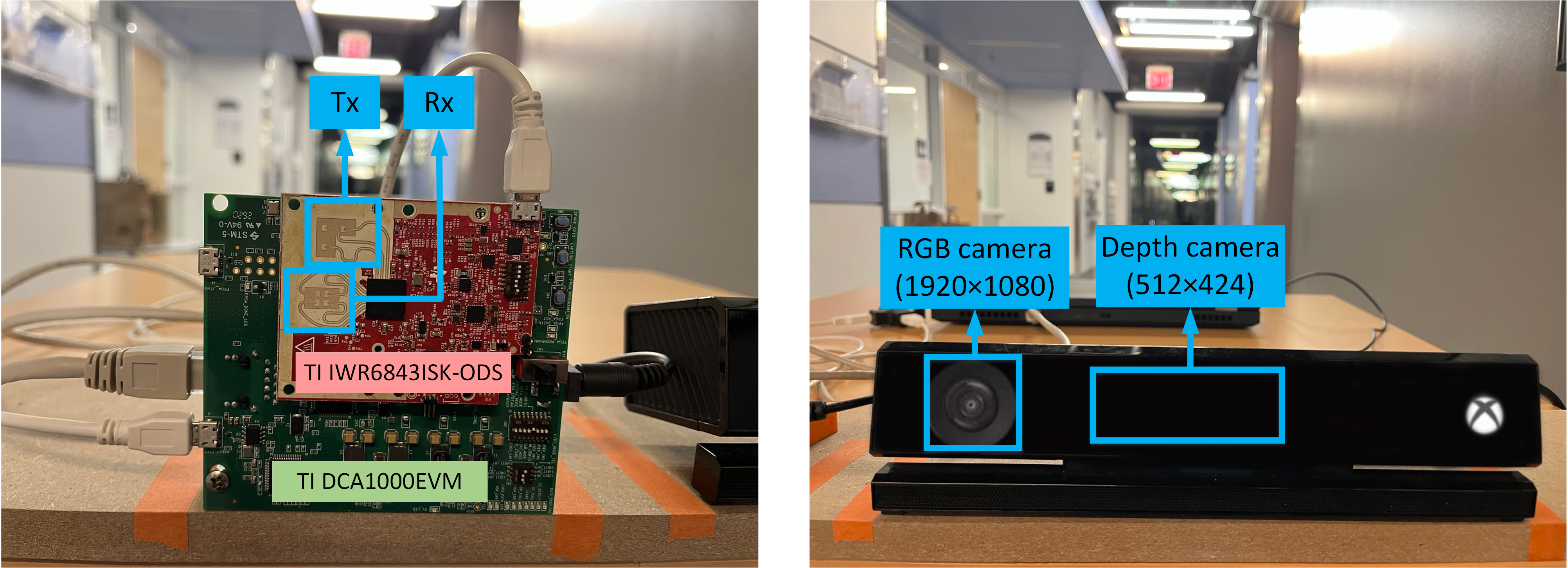}
  \caption{TI 6843ISK-ODS and Kinect V2.}
  \label{fig:sensor}
\end{figure}

As shown in Figure \ref{fig:sensor}(a), we use the TI IWR6843ISK-ODS mmWave radar for data sampling, connected to the TI DCA1000EVM board for data transmission. The radar has 3 transmitting antennas (Tx) and 4 receiving antennas (Rx). Each transmitting antenna emits its signal in an assigned time slot using MIMO technology, and the signal is received by the four receiving antennas for further processing. We set the number of chirps $C$ per mmWave frame to 64, with 256 samples per chirp. Additionally, as shown in Figure \ref{fig:sensor}(b), we use the Kinect V2 depth camera to collect depth image data. The camera operates at a 30 FPS frame rate, with a depth image resolution of 512 by 424 pixels.

\begin{figure}[htb]
  \centering
  \includegraphics[width=\linewidth]{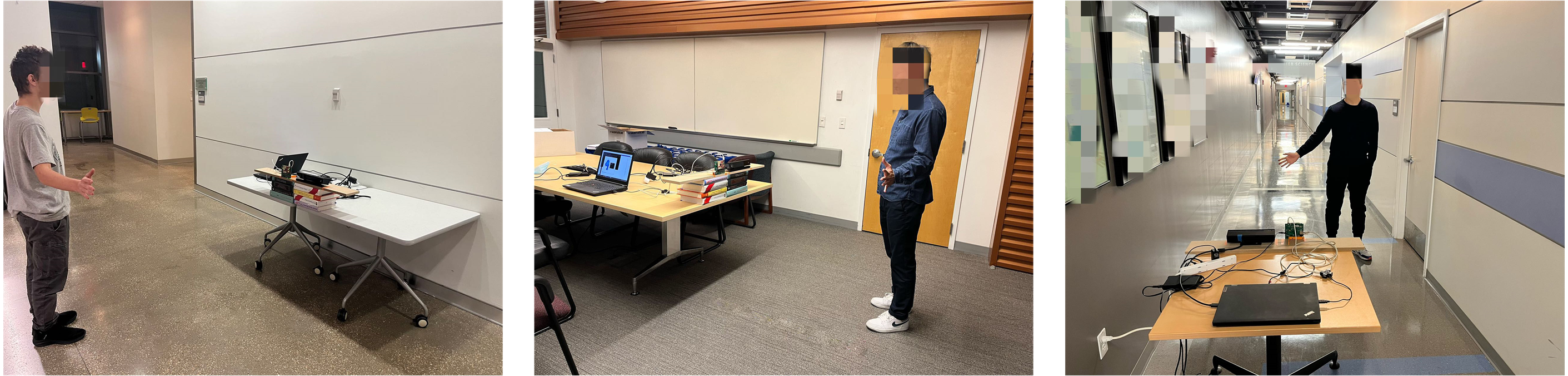}
  \caption{Three data collection scenarios: Hall, Meeting room and Corridor (from left to right).}
  \label{fig:scenario}
\end{figure}

\subsubsection{Data Collection and Labels}

As shown in Figure \ref{fig:scenario}, 12 volunteers (8 males and 4 females) are asked to move one hand freely in front of the mmWave sensor and the Kinect depth camera in three different scenarios: Hall, Meeting Room, and Corridor. These scenarios collectively represent a wide range of real-world applications for hand localization. The hall scenario covers large, open spaces such as auditoriums, testing the system's ability to handle wide-area coverage. The meeting room represents medium-sized, enclosed environments like offices, where multi-path interference from surrounding objects (e.g., chairs, tables and people) challenges the system’s hand localization in cluttered settings. The corridor scenario simulates narrow, confined spaces like hallways, where more severe environmental interference occurs. Together, these scenarios ensure the system's robustness across diverse spatial conditions.

\begin{figure}[htb]
  \centering
  \includegraphics[width=\linewidth]{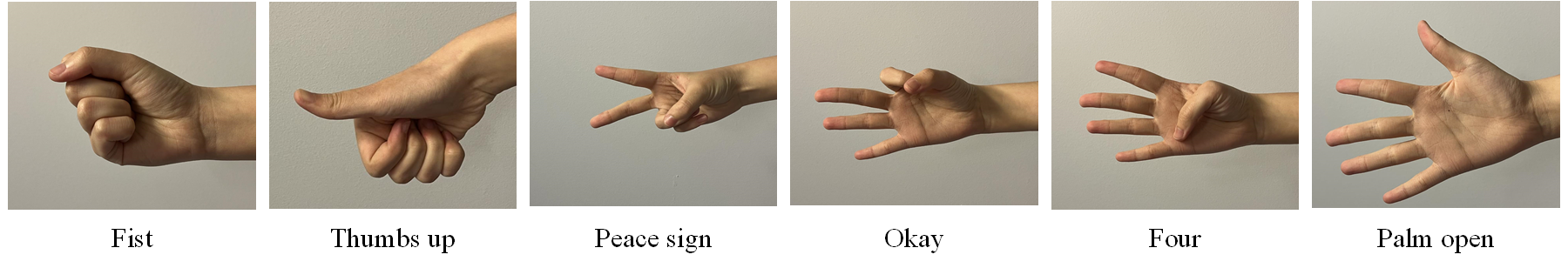}
  \caption{Six basic hand gestures.}
  \label{fig:gesture}
\end{figure}

In each scenario, we collect 5 sets of left-hand data and 5 sets of right-hand data for each subject, with each set comprising 20 seconds of hand movement. To ensure a variety of hand gestures, volunteers are asked to perform six basic hand gestures. As shown in Figure \ref{fig:gesture}, these gestures include: (1) Fist, (2) Thumbs up, (3) Peace sign, (4) Okay, (5) Four, and (6) Palm open. During data collection, subjects stand at random distances from the radar (within a 2-meter range) and move freely, adjusting their speed and transitioning between gestures. The combination of three scenarios, six gesture variations, and varying distances and movement speeds reflects the diversity and unpredictability of real-world environments, ensuring the system’s robustness and effectiveness across different conditions. We preserve the timestamps of both the mmWave data and depth images during data collection and synchronize the two data modalities based on the timestamps. The synchronized mmWave data are then converted into the mmImage format for training the hand detection model. Additionally, a pre-trained hand detection model \cite{yolov3-tiny} is used to label the mmImage in pixel space through the mmWave labeling method described in Section \ref{sec:label}. The number of subjects and synchronized labeled mmImages in each scenario are listed in Table \ref{tab:scenario}.

\begin{table}
  \centering
  \caption{The number of subjects and synchronized frames in each of the three scenarios.}
  \label{tab:scenario}
  \begin{tabular}{ccc}
    \toprule
    Scenario & Subjects & Frames number\\
    \midrule
    Hall & 12 & 4089\\
    Meeting room & 12 & 4284\\
    Corridor & 12 & 3766\\
  \bottomrule
\end{tabular}
\end{table}

\subsubsection{Model Setting} \label{sec:setting}

We use YOLOv3 as the base hand detection model due to its proven effectiveness in real-time object detection and its ability to extract robust spatial features from mmImages, outputting hand bounding boxes along with confidence values. To ensure high-quality data input, we set the structural similarity index measure (SSIM) threshold to 0.2. This threshold is determined based on the observation that compared to synchronized Kinect depth images, mmImages with an SSIM below 0.2 introduce stronger noise and degrade model performance. Only mmImages with SSIM values exceeding this threshold are included in the training process to enhance data reliability. For the loss function, the weights for cross-modality loss and label loss are both set to $\alpha _{cross}=1 $ and $\alpha _{label}=1$, respectively. These values are chosen after testing various weight combinations, where equal weighting provided the best balance between maintaining consistency across modalities and ensuring accurate label supervision. We randomly select 80$\%$ of all subjects for training, with the remaining 20$\%$ reserved for testing. This split maximizes the training data while preserving a sufficient test set to evaluate generalization. The model is trained for 100 epochs with a batch size of 8, chosen to balance GPU memory constraints and training stability. The initial learning rate is set to $10^{-3}$, and reduced to $3 \times 10^{-6}$ using a cosine annealing strategy. This learning rate schedule is chosen after comparing it with step decay and exponential decay, with cosine annealing yielding smoother convergence and improved final accuracy. The mmHand system is implemented using PyTorch, and all training is conducted on a TITAN Xp GPU. This hardware setup allows for efficient model training while maintaining flexibility for hyperparameter tuning.

\subsection{System Comparison and Performance Metrics}

\subsubsection{System Comparison}

In the experiments, we evaluate three different system structures.

\noindent
\textbf{KinectSys (Benchmark): } KinectSys is a system that includes a hand detection model trained on Kinect depth images. Due to the high resolution of these images, KinectSys achieves high accuracy in hand detection and is robust across various scenarios. It serves as the benchmark for measuring the feature extraction capabilities of other systems.

\noindent
\textbf{mmImageSys (Baseline): } mmImageSys has the same structure as KinectSys, but the hand detection model is trained on mmImage data. It serves as the baseline for evaluating the basic performance of mmImage data in hand localization.

\noindent
\textbf{mmHand system: } The mmHand system is our proposed hand localization system, sharing the same structure as KinectSys and mmImageSys, but incorporating a cross-modality connection to KinectSys to enhance feature extraction in the mmImage. The system parameter settings are outlined in Section \ref{sec:setting}.

\subsubsection{Performance Metrics} \label{sec:metric}

We use the following four metrics to evaluate the performance of our proposed hand localization system.

\noindent
$\boldsymbol{\mathcal{L}_{cross}}$ \textbf{(Cross-Modality Loss):} As introduced in Section \ref{sec:cross-loss}, $\mathcal{L}_{cross}$ measures the inverse of the feature map similarity between two hand detection models trained by mmImage and Kinect depth images, respectively. Given the high hand detection accuracy of the model trained by Kinect depth images, it excels at extracting high-resolution features. The feature map similarity evaluates the mmImage model's ability to extract similarly high-resolution features as the Kinect model.

\noindent
$\boldsymbol{\sigma ^{2}}$ \textbf{(Attention Map Variance):} Variance $\sigma^{2}$ is a metric used to measure the concentration of a model's attention map \cite{robinson2019laplace, domingos2000unified}, with higher variance indicating a more scattered attention map. Given an attention map with dimensions $(W, H)$, it can be represented as a binary function $f(x,y)$, where $x$ and $y$ are the two dimensions, with $0 \leq x \leq W$ and $0 \leq y \leq H$. The variance $\sigma^{2}$ is calculated using Equation (\ref{equ:variance}):\begin{equation}
  \sigma ^{2} =\frac{\Sigma _{x=0}^{W} \Sigma _{y=0}^{H} ((x-\bar{x})^{2}+(y-\bar{y})^{2})\cdot f(x,y)}{\Sigma _{x=0}^{W} \Sigma _{y=0}^{H}f(x,y)},
  \label{equ:variance}
\end{equation} where $\bar{x} =\frac{\Sigma _{x=0}^{W} \Sigma _{y=0}^{H} x\cdot f(x,y)}{\Sigma _{x=0}^{W} \Sigma _{y=0}^{H}f(x,y)}$ and $\bar{y} =\frac{\Sigma _{x=0}^{W} \Sigma _{y=0}^{H} y\cdot f(x,y)}{\Sigma _{x=0}^{W} \Sigma _{y=0}^{H}f(x,y)}$, and $(\bar{x}, \bar{y})$ represents the weighted center of the attention map.


\noindent
\textbf{AP (Average Precision):} AP is a commonly used metric for measuring the accuracy of object detection \cite{padilla2020survey, kim2018parallel, fang2019tinier}. It is defined as the area under the precision-recall curve, where precision and recall are calculated based on TP (True Positive), FP (False Positive), and FN (False Negative) values. To determine whether a detection is classified as TP, the IoU (Intersection Over Union) metric is used. IoU quantifies the overlap between the predicted hand bounding box and the labeled hand bounding box, and is mathematically defined as the ratio of the intersection area (overlapping region) to the union area (combined area of both boxes), given by $IoU=\frac{Area\,of\,Overlap
}{Area\,of\,Union}$. An IoU value of 1.0 indicates a perfect match, while 0.0 means no overlap. A predefined IoU threshold is set, where detections with an IoU above the threshold are classified as TP, while those below are considered FP.


\noindent
\textbf{APE (Average Pixel Error):} APE is a distance error metric used to measure the accuracy of hand localization \cite{fan2021lasot}. We calculate APE by averaging the Euclidean distance between the predicted hand center and the labeled hand center over a continuous period. Given the high accuracy of the Kinect model, we use it to provide the groundtruth hand center locations.

\subsection{System Factor Analysis}

\subsubsection{High-resolution Feature Extraction}

\begin{figure}[htb]
  \centering
  \includegraphics[width=\linewidth]{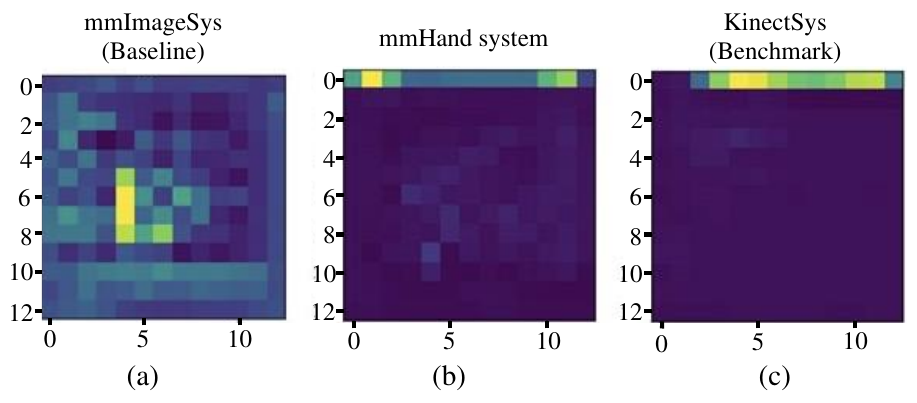}
  \caption{An example of the feature map visualization results for the three systems: (a) mmImageSys, (b) our mmHand system, and (c) KinectSys.}
  \label{fig:Feature}
\end{figure}

As shown in Figure \ref{fig:yolo}, we apply the cross-modality loss on the feature map to guide the mmImage-based model in extracting high-resolution features similar to those of the pre-trained hand detection model using Kinect depth images (i.e., KinectSys). In this section, we evaluate the effect of the cross-loss constraint in model training by comparing the average cross loss $\mathcal{L}_{cross}$ values on the testing dataset. As shown in Table \ref{tab:cross}, our mmHand system demonstrates a superior feature extraction ability compared to mmImageSys. The extracted feature map has a data format of $(K, W, H)$, where $K$, $W$, and $H$ represent the number of channels, width, and height, respectively. We average the feature map along the channel dimension $K$ and visualize the averaged feature map for comparison. As shown in Figure \ref{fig:Feature}, the feature map from the mmHand system closely resembles the feature map from the KinectSys, demonstrating the effectiveness of the cross-loss constraint in improving feature extraction.

\begin{table}
  \centering
  \caption{Comparison of high-resolution feature extraction ability between mmHand system and mmImageSys (Baseline) ($\boldsymbol{0 \le \mathcal{L}_{cross} \le 1}$, where lower $\boldsymbol{\mathcal{L}_{cross}}$ indicates better performance).}
  \label{tab:cross}
  \begin{tabular}{ccc}
    \toprule
    System Type & Scenario & $\mathcal{L}_{cross}$\\
    \midrule
    mmHand system & Hall & \textbf{0.335}\\
    mmImageSys & Hall & 1.46\\
    mmHand system & Meeting room & \textbf{0.321}\\
    mmImageSys & Meeting room & 1.73\\
    mmHand system & Corridor & \textbf{0.356}\\
    mmImageSys & Corridor & 3.27\\
  \bottomrule
\end{tabular}
\end{table}

\subsubsection{Feature Attention}

\begin{figure}[htb]
  \centering
  \includegraphics[width=\linewidth]{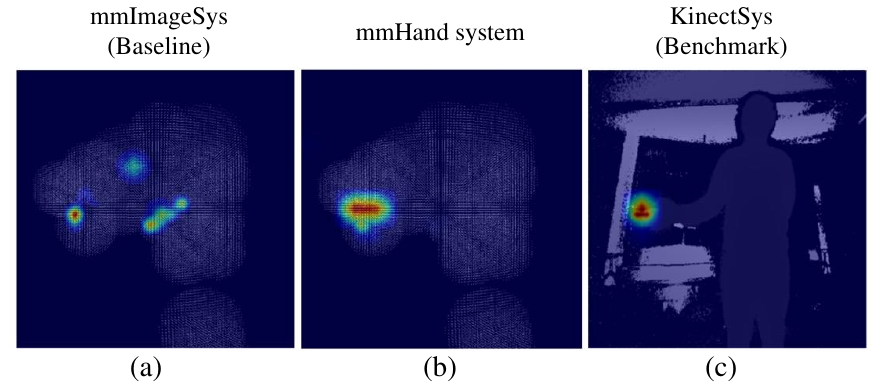}
  \caption{Visualization of the feature attention maps for the three systems: (a) mmImageSys, (b) our mmHand system, and (c) KinectSys.}
  \label{fig:Heatmap}
\end{figure}

Apart from improving high-resolution feature extraction, our cross-modality loss also guides the hand detection model to focus on specific feature areas. As shown in Table \ref{tab:attention}, we compare the feature attention ability of hand detection models in two systems (i.e., mmHand system and mmImageSys) on the testing dataset. Our mmHand system has lower $\sigma^2$ (Attention Map Variance) compared to mmImageSys, indicating that the attention map from our mmHand system is more focused and demonstrates better feature attention ability. Additionally, we visualize the feature attention maps of the two systems and KinectSys. As shown in Figure \ref{fig:Heatmap}, the visualization results further confirm the effectiveness of our cross-modality loss in improving the system's feature attention ability.

\begin{table}
  \centering
  \caption{Comparison of feature attention ability between mmHand system and mmImageSys ($\boldsymbol{0 \le \sigma^{2} \le 65025}$, where lower $\boldsymbol{\sigma^{2}}$ indicates better performance).}
  \label{tab:attention}
  \begin{tabular}{ccc}
    \toprule
    System Type & Scenario & $\sigma^{2}$\\
    \midrule
    mmHand system & Hall & \textbf{1092}\\
    mmImageSys & Hall & 1580\\
    mmHand system & Meeting room & \textbf{546}\\
    mmImageSys & Meeting room & 1837\\
    mmHand system & Corridor & \textbf{353}\\
    mmImageSys & Corridor & 683\\
  \bottomrule
\end{tabular}
\end{table}

\subsubsection{Hand Detection Results}

With improved high-resolution feature extraction and feature attention abilities, the mmImage-based model achieves more accurate hand detection. We compare the AP (Average Precision) performance of the proposed mmHand system and mmImageSys across five different IoU thresholds (ranging from 0.5 to 0.1 in steps of 0.1). As shown in Table \ref{tab:AP}, the mmHand system outperforms mmImageSys in all three scenarios, demonstrating the superiority of our cross-modality spatial-feature enhanced model for hand detection.

\begin{table*}
  \centering
  \caption{Comparison of AP (Average Precision) results between mmHand system and mmImageSys across three scenarios ($\boldsymbol{0 \le AP \le 100\%}$, where higher AP is better).}
  \label{tab:AP}
  \begin{tabular}{ccccccc}
    \toprule
    System Type & Scenario & $AP_{0.5}$ & $AP_{0.4}$ & $AP_{0.3}$ & $AP_{0.2}$ & $AP_{0.1}$\\
    \midrule
    mmHand system & Hall & \textbf{29.96\%} & \textbf{38.04\%} & \textbf{46.97\%} & \textbf{54.13\%} & \textbf{59.51\%}\\
    mmImageSys & Hall & $15.35\%$ & $21.13\%$ & $26.93\%$ & $34.69\%$ & $43.03\%$\\
    \midrule
    mmHand system & Meeting room & \textbf{21.77\%} & \textbf{33.38\%} & \textbf{43.70\%} & \textbf{57.06\%} & \textbf{67.75\%}\\
    mmImageSys & Meeting room & $15.77\%$ & $26.85\%$ & $35.18\%$ & $43.87\%$ & $56.74\%$\\
    \midrule
    mmHand system & Corridor & \textbf{21.58\%} & \textbf{30.87\%} & \textbf{46.74\%} & \textbf{57.06\%} & \textbf{70.65\%}\\
    mmImageSys & Corridor & $20.11\%$ & $28.10\%$ & $38.70\%$ & $49.28\%$ & $60.88\%$\\
  \bottomrule
\end{tabular}
\end{table*}

\subsubsection{Pixel-level Hand Localization}

\begin{figure}[htb]
  \centering
  \includegraphics[width=\linewidth]{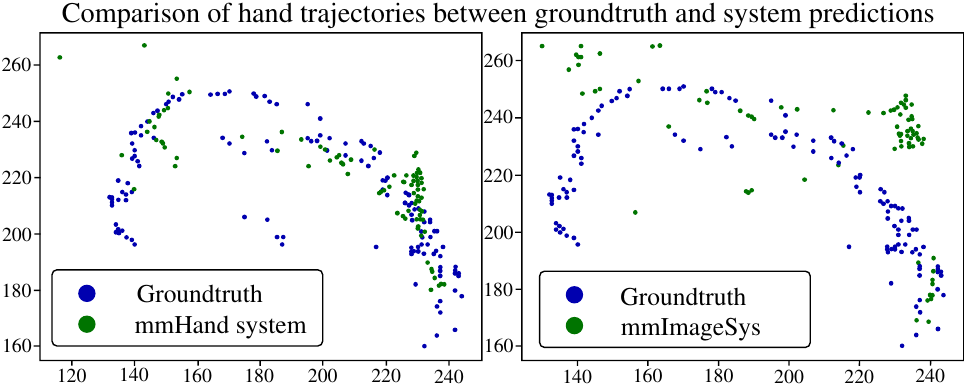}
  \caption{Example of a pixel-level hand localization result comparison. Blue points represent the groundtruth. In the left figure, green points show the predicted hand centers from the proposed mmHand system, while in the right figure, green points represent the predicted hand centers from the mmImageSys system.}
  \label{fig:trajectory}
\end{figure}

To evaluate the accuracy of our hand localization system, we sample 200 consecutive frames during six hand movement periods from each of the 12 subjects and use our mmImage-based hand localization system to predict hand locations. As discussed in Section \ref{sec:metric}, we use the predicted hand center from KinectSys as the groundtruth due to its high accuracy. We then calculate the APE (Average Pixel Error) for all subjects in each scenario. As shown in Table \ref{tab:APE}, the accuracy of our mmHand system is higher than that of mmImageSys. Additionally, Figure \ref{fig:trajectory} visualizes an example of a hand localization period, clearly demonstrating that the mmHand system achieves more accurate pixel-level dynamic hand localization.

\begin{table}
  \centering
  \caption{Comparison of APE (Average Pixel Error) results between mmHand system and mmImageSys across three scenarios ($\boldsymbol{0 \le APE \le \sqrt{2} }$, where lower APE is better).}
  \label{tab:APE}
  \begin{tabular}{ccc}
    \toprule
    System Type & Scenario & APE\\
    \midrule
    mmHand system & Hall & \textbf{0.265}\\
    mmImageSys & Hall & 0.425\\
    \midrule
    mmHand system & Meeting room & \textbf{0.146}\\
    mmImageSys & Meeting room & 0.584\\
    \midrule
    mmHand system & Corridor & \textbf{0.145}\\
    mmImageSys & Corridor & 0.157\\
  \bottomrule
\end{tabular}
\end{table}

\subsection{System Impact Factors}

\begin{figure*}[htb]
  \centering
  \includegraphics[width=\linewidth]{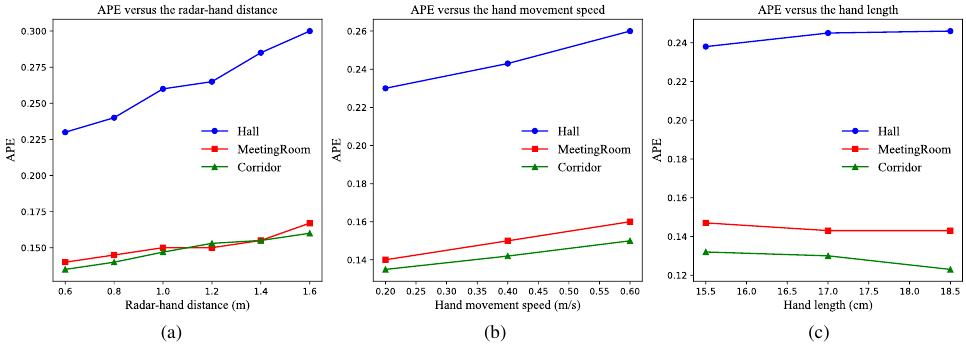}
  \caption{Experimental results for system impact factors: (a) APE (Average Pixel Error) performance versus radar-hand distance; (b) APE performance versus hand movement speed; (c) APE performance versus hand size.}
  \label{fig:factor}
\end{figure*}

\subsubsection{Radar-hand Distance}

As the distance increases, the reflected signal strength weakens, reducing detection accuracy. In this section, we evaluate the impact of radar-hand distance on hand localization performance. Specifically, the subject stands at six different distances from the mmWave sensor, ranging from 0.6 m to 1.6 m in 0.2 m increments. At each distance, we collect 200 consecutive frames over three hand movement periods, maintaining a constant movement speed of 0.2 m/s. Using the mmImage-based hand localization system to predict hand positions, we observe in Figure \ref{fig:factor}(a) that the system's APE (Average Pixel Error) increases with distance, demonstrating that shorter radar-hand distances improve localization accuracy.

\subsubsection{Hand Movement Speed}

Hand movement speed can influence the hand localization system because faster movements can lead to motion blur in the radar signal, making it harder for the system to accurately track hand positions. To evaluate this impact on the proposed mmHand system, the subjects stand 0.6 m from the mmWave sensor and move their hand at three different speeds (0.2, 0.4, and 0.6 m/s). For each speed, we collect 200 consecutive frames over three hand movement periods to test the system's localization performance. As shown in Figure \ref{fig:factor}(b), the APE increases with movement speed, demonstrating that faster hand movement reduces localization accuracy.

\subsubsection{Hand Size}

Theoretically, hand size can influence the hand localization system, as larger hands reflect more radar signals, resulting in a stronger and clearer signal for the system to process. In contrast, smaller hands may produce weaker reflections, making accurate detection and localization more challenging. However, since the range of hand size variation is relatively small \cite{CPWR}, its overall impact is minimal. Typically, hand size is measured by hand length, defined as the distance from the tip of the middle finger to the base of the hand \cite{Ansell}. To assess the impact of hand size on our mmHand system, we collect hand movement data from three subjects with different hand lengths (15.5, 17, and 18.5 cm). For each hand length, we record 200 consecutive frames over three hand movement periods to evaluate the system's localization performance. As shown in Figure \ref{fig:factor}(c), the APE performance remains relatively consistent, indicating that hand size has little effect on the hand localization performance of the mmHand system.

Based on the experimental results of the three system impact factors, although radar-hand distance and hand movement speed influence system performance to some extent, the low APE ensures that the mmHand system remains viable for many real-world applications \cite{li2024beyond, ettu2024analysis, schneider2021accuracy}. 

\section{Discussions}

\subsection{System Latency Analysis}

We use the Intel Xeon CPU E5-1620 v4 @ 3.5GHz to process the mmWave sensor data, while an Nvidia Titan XP is deployed for hand localization model predictions. The data format for each collected frame is set as $(\mathcal{T}, \mathcal{R}, \mathcal{C}, \mathcal{N})=(3, 4, 64, 256)$. The time cost for each component of the mmHand system is presented in Table \ref{tab:latency}, with the total system latency being approximately 0.1 seconds, which is sufficient for most real-time applications. Additionally, the data processing component of the system can be further accelerated in two areas.

First, the number of chirps per frame $\mathcal{C}$ is related to the velocity resolution. Reducing $\mathcal{C}$ will not affect the spatial resolution of the mmWave sensor, but it can decrease the mmWave data size and accelerate mmImage generation. However, too few chirps can lead to instability in mmImage generation, so balancing $\mathcal{C}$ is essential for achieving real-time performance in the mmHand system while maintaining localization accuracy. Second, in the mmImage generation tool, MVDR generates range-azimuth-elevation data in the format $(\mathcal{D}, \Theta, \Phi)$, where $\mathcal{D}$ ranges up to 10 meters. In practice, if the human object is too far from the mmWave sensor, the reflected signal will have low energy and be heavily affected by environmental noise. Additionally, several works have demonstrated coarse-grained localization of objects from the range bin $(\mathcal{T}, \mathcal{R}, \mathcal{C}, \mathcal{D})$, which serves as the input to MVDR \cite{bhatia2021classification, hamidi2021single, zhao2018millimeter}. Therefore, it is feasible to add an object range detection module to filter out data outside the object's distance range, significantly reducing the time cost of MVDR.

\begin{table}[htb]
\small
  \centering
  \caption{mmHand system latency analysis.}
  \label{tab:latency}
  \begin{tabular}{ccc}
    \toprule
    Data collection & Data processing & Model detection\\
    \midrule
    0.03s & 0.05s & 0.02s\\
  \bottomrule
\end{tabular}
\end{table}

\subsection{mmHand for Future Applications}

\noindent
\textbf{Multi-hand Localization:} Currently, our current mmHand system is capable of dynamically localizing a single hand with various gestures. However, with advancements in resolution and feature attention enabled by cross-modality learning, there is potential to classify and localize multiple hands from the generated mmImages \cite{xue2022m4esh, wu2020mmtrack}. In future work, we plan to explore multi-hand localization using the mmHand system in simple scenarios, such as two hands from one or two individuals moving without overlap.

\noindent
\textbf{Hand Pose Recognition:} Since the generated mmImage has high spatial information similarity with the synchronized depth image, it is possible to extract coarse-grained hand contour features from the mmImage, which can be used for hand pose recognition. To date, many super-resolution methods have been developed to improve object recognition accuracy \cite{liu2021accurate, zhang2018image, anwar2020densely}. In the future, we plan to integrate these methods into our mmHand system to achieve accurate hand pose recognition.

\noindent
\textbf{Occlusions:} Due to the inherent penetration ability of mmWave signals \cite{pegoraro2022oracle, xue2021mmmesh}, mmWave sensor can detect objects behind occlusions, whereas vision-based cameras lose functionality in such conditions. However, occlusion causes significant signal attenuation, making the received signal more vulnerable to environmental interference. Additionally, the material structure of the occlusion can induce phase shifts in the received signal, both of which negatively impact mmWave data quality. In the future, we will further investigate how different occlusions affect data quality and apply our mmHand system for hand localization in occluded environments.

\subsection{mmHand for Industry Applications}

In addition to future research directions, the mmHand system has significant potential for industrial applications, particularly in environments where traditional camera-based systems face limitations. Below, we discuss three example use cases:

\noindent
\textbf{Quality Inspection in Manufacturing:} Camera-based systems in the manufacturing operation line often fail when lenses are obscured by dust or grease. For instance, in food processing plants, workers frequently interact with machinery to adjust settings or sort items on a conveyor belt. Over time, airborne grease particles or fine food dust can accumulate on camera lenses, significantly blocking the camera from detecting and localizing hands accurately \cite{kold2016conveyors, kohli2012methods}. This can lead to system failures in detecting wrong operations during production, thus making the product of bad quality. The deep learning approach can mitigate the impact of interfering substances to some extent through data-driven model training \cite{gutierrez2021synthetic, kim2021product, zheng2021recent}. However, the complex and dynamic nature of the manufacturing environment, along with the diverse sources of interference, poses significant challenges for model training \cite{xu2022review, yang2020using, malhan2023role}. In contrast, the mmHand system operates using the mmWave signal to penetrate these contaminants, ensuring reliable hand localization. This robustness reduces downtime for cleaning and maintenance, making mmHand particularly suitable for continuous operation in contaminated environments \cite{zekri2020analysis, ryan2017indoor}. Additionally, the privacy-preserving nature of mmWave signals ensures that sensitive operation processes remain secure \cite{sun2019privacy}.

\noindent
\textbf{Sterile Environments in Healthcare:} In healthcare, particularly in operating rooms or intensive care units, maintaining sterility is critical \cite{nazhat2020sterility, ma2012rfid}. For example, during a surgical procedure, surgeons often need to interact with medical devices without physically touching them to avoid contamination. Camera-based systems in such settings may fail due to condensation on the lens from the operating room's humid environment or blood splatters obstructing the camera's view \cite{mohsan2021optical, scott2022healthcare}. The mmHand system, however, operates effectively under these conditions because mmWave signals are unaffected by visual obstructions. Moreover, the mmHand system’s penetration capability allows it to function even through thin surgical drapes or plastic covers, which are commonly used to maintain sterility \cite{hussain2019different, sun2022human}. Additionally, mmWave radiation operates within the non-ionizing spectrum, with energy levels significantly lower than those of X-rays or CT imaging systems, posing no known health risks to patients or medical staff \cite{dilli2021implications}. These attributes make mmHand a safe and effective solution for enhancing precision and efficiency in sterile surgical environments.

\noindent
\textbf{Harsh Conditions in Mining or Heavy Industries:} Mining operations or heavy industries present challenging conditions, including high levels of dust, vibration, and debris \cite{cecala2012dust, colinet2021best, liu2020impacts}. In these environments, the ability of mmHand to penetrate through obstructions and maintain accuracy makes it a reliable tool for remote control of heavy machinery \cite{singh2019radhar}. With its compact form factor (millimeter-level antenna size) and low power consumption (within 5 mW \cite{IWRL6432}), the mmHand system is well-suited for integration into wearable edge devices that enable real-time control of machinery in mining environments. For instance, in underground mining, workers often operate machinery in low-visibility conditions due to airborne dust particles. The camera-based hand localization system doesn't work as the lens becomes obscured by dust \cite{yaghoobi2017vehicle}. In contrast, the mmHand system remains unaffected by such obstructions, ensuring accurate hand tracking and control of machinery.


\section{Related work}

\noindent
\textbf{Cross-modality Learning:} Cross-modality learning is an effective method to combine the advantages of different data modalities, leading to improved performance. Currently, fusion and guidance are the two primary approaches in cross-modality learning technologies \cite{chen2019crdoco, li2019attention, song2021tacr, wang2020dual, thoker2021feature, sun2021learning}. In the fusion approach, features extracted from different data modalities are combined to enhance the performance of specific model tasks \cite{bruce2021multimodal, zhang2021deep, debie2019multimodal, lin2020ensemble}. For example, Xue et al. \cite{xue2019deepfusion} introduce a DeepFusion model that integrates features from different sensor modalities along with cross-sensor correlations, improving performance in IoT applications. Shuai et al. \cite{shuai2021millieye} propose a lightweight mmWave sensor and camera fusion system for more robust object detection, using mmWave sensor to address lighting issues and RGB images to enhance model accuracy. In contrast, the guidance approach enables a high-quality data modality to guide feature extraction from a lower-quality modality \cite{romero2014fitnets, gou2021knowledge, beyer2022knowledge}. Cai et al. \cite{cai20203d} develop a cross-modality interaction between depth images and RGB images, improving the encoder's ability to extract more precise depth information from RGB data. Zhao et al. \cite{zhao2018through} use RGB data to guide RF signal model training, effectively improving human pose estimation from RF signals. Unlike these works, our mmHand system explores a more efficient cross-modality learning approach by investigating a new image-format mmWave data representation for deeper integration between depth images and mmWave data.

\noindent
\textbf{Hand Localization Sensor Systems:} Hand localization has been a widely studied problem for several years, with various technologies explored to achieve accurate localization using different data modalities \cite{cui2021hand, ohkawa2022domain, che2019novel, shimizu2021hand, koitka2018ossification, sun2022human}. For example, Liu et al. \cite{liu2015rgbd} present a dynamic hand localization and gesture recognition system based on RGBD video, which provides rich visual information but raises privacy concerns and requires sufficient and stable lighting conditions. Baldi et al. \cite{baldi2015using} develop a sensing glove with wearable sensors such as IMU, ECG, and EMG, which ensures accurate tracking but may cause user discomfort due to sensor placement and battery limitations. Wang et al. \cite{wang2018witrace} propose WiTrace, achieving centimeter-level passive hand localization using WiFi signals, though it necessitates dense WiFi deployments. Chen et al. \cite{chen2017echotrack} implement a smartphone-based prototype using acoustic signals to locate hand positions, which avoids occlusion issues but is susceptible to environmental sound pollution. Additionally, McIntosh et al. \cite{mcintosh2017improving} utilize ultrasonic waves for hand tracking, offering an alternative approach but facing challenges related to high costs and large device sizes. Table \ref{tab:modality} highlights the advantages of mmWave data compared to other data modalities \cite{hussain2019different, zhao2022human}. Unlike existing hand localization technologies that rely on specific sensor placements or complex data processing algorithms, our mmHand system provides privacy-preserving, device-free hand localization without these requirements.

\begin{table*}[htb]
\small
  \centering
  \caption{Advantages of mmWave data compared to other data modalities.}
  \label{tab:modality}
  \begin{tabular}{ccc}
    \toprule
\textbf{Data modality} & \textbf{Technology} & \textbf{Advantages of the mmWave modality}\\
    \midrule
    Radio signal & WiFi, RFID and Bluetooth & Higher resolution\\
    \midrule
    Acoustic signal & Microphone and Speaker & Not affected by sound pollution\\
    \midrule
    Camera image & RGB and depth image & Privacy-preserving and light-free\\
    \midrule
    Ultrasound & Ultrasonic & Lower price and smaller size\\
    \midrule
    Wearable sensor data & IMU, ECG and EMG & \makecell[c]{No accumulated error, better user comfort \\and no battery capacity limitation}\\
  \bottomrule
\end{tabular}
\end{table*}

\section{Conclusion}

In this paper, we introduce mmHand, a novel hand localization system that achieves pixel-level accuracy using a single commodity mmWave radar. The system requires no specific sensor placement and can accurately predict dynamic hand locations in pixel space. A new mmImage generation tool is designed to fully extract spatial information from raw mmWave data and represent it in pixel space. Additionally, the system introduces innovative methods for quality evaluation and pixel space labeling of time-series mmWave data. Leveraging the spatial information similarity between mmWave data and camera depth images, the system also features a cross-modality spatial-feature enhanced model for more accurate pixel-level hand localization. Experiments with 12 subjects across 3 different scenarios, using 4 metrics, demonstrate the efficiency of our mmHand system in hand localization.

\section{Acknowledge}
This work is partially supported by US NSF Award $\#$2426470 and the 2022 Meta Research Award

\bibliographystyle{IEEEtran} 
\bibliography{reference}


\end{document}